\documentclass{article}
\usepackage{ijcai21}

\usepackage{times}
\usepackage{soul}
\usepackage{url}
\usepackage[hidelinks]{hyperref}
\usepackage[utf8]{inputenc}
\usepackage[small]{caption}
\usepackage{graphicx}
\usepackage{amsmath}
\usepackage{amsthm}
\usepackage{booktabs}
\usepackage{algorithm}
\usepackage{algorithmic}
\usepackage{amsfonts,dsfont}  
\usepackage{mathtools, xparse}
\usepackage{nicefrac}         
\usepackage{subfig}
\usepackage{float}
\usepackage{enumitem}
\usepackage{array}
\usepackage{lscape}

\newtheorem{theorem}{Theorem}
\newtheorem{definition}{Definition}[section]
\newtheorem{theorem lemma}{Theorem}
\newtheorem{lemma}[theorem lemma]{Lemma}

\newcommand{\model}{G-FORCE~}
\newcommand{\groupaware}{GroupAware~}
\newcommand{\eo}{equalized odds~}
\newcommand{\eons}{equalized odds}

\makeatletter
\def\blfootnote{\gdef\@thefnmark{}\@footnotetext}
\makeatother

\DeclarePairedDelimiter\abs\lvert\rvert

\title{Towards Reducing Biases in Combining Multiple Experts Online}

\author{
    Yi Sun$^1$ 
    \and
    Ivan Ramirez$^1$ \and
    Alfredo Cuesta-Infante$^2$   \And
    Kalyan Veeramachaneni$^1$
\affiliations
$^1$ MIT
$^2$Univ. Rey Juan Carlos\\
\emails
\{yis, iramdia\}@mit.edu,
alfredo.cuesta@urjc.es,
kalyan@csail.mit.edu
}

\begin{document}

\maketitle

\begin{abstract}
  In many real life situations, including job and loan applications, gatekeepers must make justified and fair real-time decisions about a person’s fitness for a particular opportunity. In this paper, we aim to accomplish approximate group fairness in an online stochastic decision-making process, where the fairness metric we consider is \eons. Our work follows from the classical learning-from-experts scheme, assuming a finite set of classifiers (human experts, rules, options, etc) that cannot be modified. We run separate instances of the algorithm for each label class as well as sensitive groups, where the probability of choosing each instance is optimized for both fairness and regret. Our theoretical results show that approximately equalized odds can be achieved without sacrificing much regret. We also demonstrate the performance of the algorithm on real data sets commonly used by the fairness community.
\end{abstract}


\section{Introduction}
In the near future, machine learning models are expected to aid decision-making in a variety of fields. Recently, however, concerns have arisen that machine learning models may accentuate preexisting human biases and data imbalances, affecting decision-making and leading to unfair outcomes in areas including policing \cite{COMPAS_2016}, college admissions, and loan approvals. 

The machine learning community has responded to this critique by designing strategies to train a fairer classifier \cite{zafar2015fairness}\cite{Donini2018}, yet this is not always a feasible solution. 
There are many situations in which decisions are not made by a trainable classifier, but instead by human experts or black-box models.
A better solution, then, would be an online policy balancing accuracy and fairness without considering individual experts’ technical underpinnings. 
Indeed, ensuring fairness \textit{via} a mathematical framework would ideally not only prevent prejudice within algorithms, but help quantitatively overcome human biases as well.

Here another important question arises: what, exactly, is fairness? In general, the machine-learning community agrees to aim for statistical parity -- the equalization of some measure of errors across protected groups. 
These measures include equal opportunity \cite{Hardt_2016}, \eo (disparate mistreatment) \cite{Zafar_2017}\cite{Hardt_2016} and predictive parity \cite{Celis2019}.
In this paper, we focus on achieving \eo and emphasizing that all error types should be addressed, i.e., equalized across groups, rather than equalizing a combined measure. Previous online strategies for fairness do not solve for \eo \cite{Blum_2018}\cite{Bechavod_2019}.  

Finally, in many cases, people will strategically react to a decision-making process, leaving models to face shifting distributions. For instance, if a particular job has historically employed more males than females, the position might attract more males to apply, worsening the bias in the distribution. In this situation, having fair base models and a static combining strategy is not enough to overcome bias. Even without access to the base models, an online algorithm must be adaptive over time and maintain equalized odds.

Thus this paper focuses on: an online setting where we have protected underrepresented groups; each individual is randomly sampled \textit{i.i.d.} from a potentially biased and shifting distribution across groups and labels; and a finite set of classifiers (or experts) that have already been trained where we only have access to their decisions. The goal is to design a randomized algorithm that combines these experts such that the violation of fairness of the resulting ``\textit{combined expert}'' can be upper bounded, which further allows us to achieve an optimal balance between fairness and regret.

In this paper we make the following contributions.
We propose a randomized algorithm that
~(i) works in online stochastic settings (individuals sampled from \textit{i.i.d.} distribution),
~(ii) has a provable asymptotic upper bound on regret and \eons,
~(iii)  the upper bound is minimized to achieve the desired balance between equalized positive rate, equalized negative rate and regret. These give the name \model (Group-Fair, Optimal, Randomized Combination of Experts).
Finally, we demonstrate its performance on datasets commonly used by the fairness community, and on synthetic datasets to test its performance under extreme scenarios.

\section{Related Work}\label{related}
There are two broad definitions for fairness: individual fairness and group fairness. Individual fairness builds upon ``treating similar individuals similarly'' \cite{Dwork2012}. 
On the other hand, group fairness is achieved by balancing certain statistical metrics approximately across different demographic groups (such as groups divided by gender, race, etc). Equalized odds \cite{Zafar_2017} \cite{Hardt_2016}, a.k.a. disparate mistreatment, requires that no error type appears disproportionately for any one or more groups.  A weaker notion of \eo is equal opportunity, which aims to achieve equal false positive rates \cite{Hardt_2016}.
\eo can be achieved by adding in additional constraints when optimizing the objective function \cite{Zafar_2017}, or by enforcing an optimal threshold for a predictor during post-processing \cite{Hardt_2016}. 
However, recent work shows that it is impossible to achieve \eo \cite{Chouldechova_2017,kleinberg_2017} simultaneously with other notions of fairness such as calibration, which requires that outcomes are independent of protected attributes conditional on the estimates from predictors. For example, among loan applicants estimated to have a 10\% chance of
default, calibration requires that whites and blacks actually default at similar rates.
 
In this paper we consider \eo as a fairness metric, but the method can be developed to optimize for other fairness metrics too.  It is also generally accepted that there is often a trade-off between predictive accuracy and fairness \cite{CorbettDavies2017AlgorithmicDM}.

At the same time, there has been recent interest in studying fairness in an online setting, particularly the bandit setting where the goal is to fairly select from a set of individuals at each round.  \cite{Joseph2016FairnessIL} studies the contextual bandit setting with unknown reward functions. \cite{Gillen_2018} considers a problem where the specific individual fairness metric is unknown.  \cite{Liu2017} considers satisfying calibrated fairness in a bandit setting.
On the group fairness side, \cite{Bechavod_2019} tries to enforce the equalized opportunity constraint at every round under a partial feedback setting, where only true labels of positively classified instances are observed. 
\cite{Blum_2018}, specifically shows that it is impossible to achieve \eo under an adversarial setting when an adaptive adversary can choose which individual to arrive. Our paper considers a more realistic stochastic setting, and proposes an algorithm can achieve approximate \eo with a slight increase in regret.

\section{Setting and preliminaries}\label{setting}

In this section, we introduce our setting and the notation we use throughout this paper.

\textbf{Setting.\quad} 
We assume access to a set of black-box classifiers $\mathcal{F}=\{f_1,\ldots,f_d\}$.
While many fairness solutions attempt to optimize a classifier, our goal is to produce fair and accurate predictions by combining the classifiers according to their past performances.
In online settings, an algorithm runs through rounds $t=1,\ldots,T$. At each round $t$:
\begin{itemize}[noitemsep,topsep=0pt,parsep=0pt,partopsep=0pt]
    \item  A single example $(x^t,z^t) \in \mathbb{R}^n$ arrives, where $x^t\in X$ is a vector of unprotected attributes and $z^t \in Z$ is a protected or sensitive attribute.
    \item One classifier $f^t \in \mathcal{F}$ is randomly selected to estimate $\hat{y}^t = f^t(x^t,z^t)$, the label for the input example.
    \item The true label $y^t$ is revealed after the prediction.
\end{itemize}

\textbf{Notation and assumptions.\quad} 
In this paper we consider binary classification problems, with a positive and a negative class, i.e., $y \in \{+,-\}$. For the sake of notation, we assume the whole data set can be divided in two population groups $A$ and $B$ according to the sensitive attribute; i.e.  $z \in \{A,B\}$, though our approach can be easily extended to multiple groups.
Superscript $t$ denotes the time index or \textit{round} $t$; for instance $y^t$ is the true class of the example given in round $t$.
Superscript $*$ denotes optimality; for instance $f^*(z,y)$ represents the best classifier on group $z$ with label $y$. 
Superscript $T$ denotes the matrix transpose only when it is on a vector. 
Letter $\ell$ denotes a loss function.

Throughout this paper it is often necessary to refer to a classifier $f$, a group $z$, the true label $y$, or a combination of these. We indicate such a combination with a list of subscripts; e.g. $w_{f,z}$ is the multiplicative weight associated to a given classifier $f$, specific to samples from group $z$,  
and $\ell_{f,z,y}$ represents the loss specific to samples from group $z$ and with true label $y$.
These subscripts are replaced with a specific value when necessary. For instance $\ell_{f,z,-}$ specifies that all the samples considered are negative. A lack of subscripts represents the generic variable.
We denote the probability an example coming from group $z$ as $p_z$; and define the \textit{base rates for outcomes in group $z$} as 
$\mu_{z,+} = \mathbb{P}(y=+|z)$.

\textbf{Performance metric.\quad} 
A frequent performance metric in online learning is \textit{Regret}, which compares the performance of the algorithm to that of the best fixed expert in hindsight.
\vspace*{-0.05in}
\[ \label{eq:Regret}
    \mathrm{Regret}(T)  = \sum_{t=1}^T\ell(f^t(x^t,z^t),y^t) 
    - \inf_{f \in \mathcal{F}}\sum_{t=1}^T\ell(f(x^t,z^t),y^t)
\]
The typical goal of online learning algorithm is to achieve sub-linear regret over $T$ rounds; i.e. $  \lim\limits_{T \to \infty} {\mathrm{Regret}(T)} / {T} = 0. $
In addition, we try to make the algorithm fair to each sensitive group by considering \eo:

\begin{definition}[Equalized FPR and FNR] 
Let $\hat{y}$ be the estimated outcome from a binary classifier when it receives an instance with protected attributes $z$ and ground truth $y$.
\[\text{Let} \hspace{1em}\mathrm{FPR}_z = \mathbb{P}(\hat{y}=+|z,y=-)
\hspace{1em} \text{and} 
\]
\[
\mathrm{FNR}_z = \mathbb{P}(\hat{y}=-|z,y=+)
\]

be the False Positive Rate (FPR) and the False Negative Rate (FNR) for group $z$. A classifier satisfies Equalized FPR (FNR) on group A and B if \hspace{0.3em}$ \mathrm{FPR}_A = \mathrm{FPR}_B  $ ($\mathrm{FNR}_A = \mathrm{FNR}_B  $).
\end{definition}

\begin{definition}[Equalized odds]\label{def:eo}
A classifier exhibits equalized odds if it achieves both an equalized FPR and an equalized FNR.
\end{definition}
\vspace{-0.5em}
In other words, \eo implies that the outcome of the classifier is independent of the protected attributes, given the true class. 

\begin{definition}[Equalized Error Rates] 
A classifier satisfies equalized error rates (EER) on group $A$ and $B$ if 
\[\mathbb{P}(\hat{y} \neq y|z=A) = \mathbb{P}(\hat{y} \neq y|z=B)\]

\end{definition}

\begin{definition}[$\epsilon$-fairness]
 A randomized algorithm satisfies $\epsilon$-fairness if:
 $$ \small
 \vert \mathrm{FPR}_A - \mathrm{FPR}_B \vert \leq \epsilon 
 {\it~~~~~ and ~~~~} 
 \vert \mathrm{FNR}_A - \mathrm{FNR}_B \vert \leq \epsilon.
 $$

\end{definition}

Thus, a measure of how an algorithm performs in terms of \eo is given by:  $\vert \mathrm{FPR}_A - \mathrm{FPR}_B \vert$ and $\vert \mathrm{FNR}_A - \mathrm{FNR}_B \vert
 $, which are sometimes referred throughout the paper as equalized FPR or equalized FNR for simplicity.

\label{sec:FaiRMW}

\vspace{-1mm}
\paragraph{Multiplicative Weights Algorithm}
The \textit{Multiplicative Weights }(MW) is a well-known algorithm for achieving no regret by combining decisions from multiple experts.
The main idea is that the decision maker maintains weights on the experts based on their performances up to the current round, and the algorithm selects an expert according to these weights.

\begin{theorem}
\label{thm:mw_regret}
Assume that the loss $\ell_{f}^t$ is bounded in [0,1] and $\eta<\frac{1}{2}$. 
Let $\ell_{f^*}^t$ be the loss of the best expert after t rounds, then we have:
$$\sum_{t=1}^T  \pi^t \ell^t \leq (1+\eta)\sum_{t=1}^T \ell_{f^*}^t+\frac{\ln d}{\eta}, $$
where $\pi^t$ is the probability mass function (PMF) for selecting the set of classifiers at each round.
\end{theorem}

This powerful theorem \cite{Arora2012TheMW} shows that the expected cumulative loss achieved by the MW algorithm is upper bounded by the cumulative loss of the best fixed expert in hindsight asymptotically. In other words, the MW algorithm achieves sub-linear regret.

\vspace{-1mm}
\paragraph{\groupaware algorithm}

\cite{Blum_2018} proposed a \groupaware version of the MW algorithm to achieve equalized error rates (EER) across sensitive groups in an adversarial setting. 
They showed that to attain EER it is necessary to run separate instances of the original MW algorithm on each sensitive group. 
One drawback of the \groupaware algorithm is that it only bounds the performance of the overall error for each sensitive group, without any guarantee on the number of false positives and false negatives within each group. In the worst case, \groupaware could have 100\% FPR on one sensitive group and 0\% FPR on the other, which leads to a severe violation of \eons.

\section{\model~algorithm}\label{sec:algorithm}

We argue that it is necessary to run MW instance on a more granular level, in order to satisfy \eons. Specifically, it is necessary to keep different instances of the algorithm for each group as well as for each label. 

We propose a novel randomized MW algorithm that utilizes not only sensitive groups but also their labels,
and show that it is possible to find bounds on the violation of \eons. Moreover, the bounds can be further optimized by cleverly coordinating between the instances.
For the sake of clarity, the rest of the paper we consider binary classification with two sensitive groups, though the algorithm can be easily extended to multi-group and multi-class problems.

\subsection{\model mechanism}

\begin{figure*}[htp]
    \centering
    \includegraphics[scale=1.8]{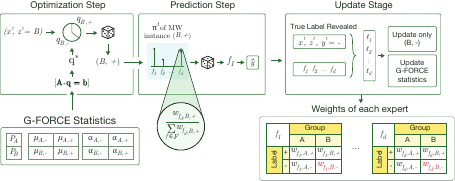}
    \caption{This figure shows how \model process an input pair $(x,z)$, where $z$ assumed to be B. In the optimization step, \model samples from PMF $[q_{B,+},q_{B,-}]$ constructed from \model statistics and selects MW instance (B,+) to use. In prediction step, instance (B,+) samples a classifier $f_1$ to predict. In the update stage, the true label revealed to be $-$, indicating that \model selected the wrong instance to use in the first stage. \model only updates the weights for the correct instance (B,-), as well as the \model statistics.  }
    \label{fig:process}
\end{figure*}

\model keeps separate MW instances for each possible 2-tuple $(z,y)$ with $z\in\{A,B\}$ and $y\in\{+,-\}$.   Throughout the paper, we uses tuple $(z,y)$ to refer to a MW instance trained for subset of data with group $z$ and label $y$.
Each MW instance associates a weight to a classifier $f$ for group $z$ and label $y$; e.g. the weight of classifier $f$ for group $A$ and negative label examples is denoted as $w_{f,A,-}$. The mechanism of \model is explained in Figure \ref{fig:process}. At each round, \model takes in an example $(x, z)$.\hspace{1em}\model works in three steps: \texttt{optimization step},  \texttt{prediction step} and the \texttt{update step}. 

\paragraph{Optimization Step} \model first selects an appropriate MW instance to use. While group $z$ is known, at this point \model doesn't know the label yet (which is exactly the target to predict) and has to choose between instance $(z,+)$ and instance $(z,-)$. \model constructs a meta probability, which we refer to as the \textbf{blind selection rule} to select between the two instances, where $q_{z,+}$ and  $q_{z,-}$ are the probability of selecting $(z,+)$, and  $(z,-)$ respectively. In the case \model selects the wrong instance (for example, true label is $-$ but $(z,+)$ is selected), we refer to the additional losses as Cross-Instance Cost $
\alpha_{z,+}$ (formal definition in next section). This meta probability allows us to explicitly construct a upper bound on the \eons. We later show $q_{z,+}$ and $q_{z,-}$ can be explicitly set to tighten this bound by solving a linear system $\textbf{Aq = b}$. The parameters of the system ($\textbf{A,b}$) depend on statistics $p_z, \mu_{z,y},\alpha_{z,y}$,which can all be estimated on the fly. We refer to these statistics as \model Statistics, and the solution of the linear system as $\textbf{q}^*$.

\paragraph{Prediction Step} Suppose instance $(z,+)$ is selected, \model uses normalized weights $\pi_{f,z,+} = \frac{w_{f,z,+}}{\sum_f w_{f,z,+}}$ to sample a classifier $f$, and adopts $f$'s prediction for this round. 

\paragraph{Update Step} After the prediction, the true label $y$ is revealed and each classifier $f$ produces loss $\ell_{f,z,y}^t = \ell(f(x,z),y)$.
\model only updates the weights for instance $(z,y)$ with the exponential rule
\[
w_{f,z,y}^{t+1} = w_{f,z,y}^{t}\left(1-\eta\right)^{\ell_{f,z,y}^t}.
\]
In addition, we also update the \model Statistics used to compute $\textbf{q}^*$. Note that although we recalculate $\textbf{q}^*$ at early rounds since estimation of \model Statistics has not converged, as the estimation of \model Statistics converge to the true value, $\textbf{q}^*$ would also converge.

\subsection{Theoretical Analysis of \model}

One key contribution of this paper is to show that: (1) the fairness loss in \model can be asymptotically upper bounded as a function of $q_{z,+}$ and $q_{z,-}$,and (2) the function values can be reduced to zeros by solving for $q_{z,+}$ and $q_{z,-}$, which further minimizes the upper bound.
Specifically, 
$$|FPR_A - FPR_B| \leq |G_{FPR} + Q_{FPR}|$$  
$$|FNR_A - FNR_B| \leq |G_{FNR} + Q_{FNR}|$$
where $G_{FPR}, G_{FNR}$ are constants that depend on the factors intrinsic to the problem (data distribution and the underlying metrics of the experts), and $Q_{FPR}, Q_{FNR}$ are functions of  $q_{z,+}$ and $q_{z,-}$. A formal version of the theorem is stated in \ref{theorem:fairness}.

In this section, we aim to develop an upper bound on the fairness loss for G-FORCE. 
We start by first providing an upper bound on regret for the worst cases scenarios, as well as a lower bound on regret for the best case scenarios (we leave the proof to the appendix).

Since there is randomness involved in the selection of MW instances, we define the costs of using a sub-optimal instance as cross-instances cost.
\begin{definition}[Cross-instances cost] 
Let $\pi_{f,z,y} =  \frac{ w_{f,z,y}^t}{  \sum_{f}w_{f,z,y}^t }$ denote the normalized weights when choosing instances (z,y). We define the cross-instances cost at round t as: 
$$\alpha^{t}_{z,y'} =  \displaystyle \underbrace{\sum_{f \in \mathcal{F}}\pi_{f,z,y'} \cdot \ell_{f,z,y}^t}_{\text{expected losses with wrong instance (z,y')}} - \underbrace{\sum_{f \in \mathcal{F}} \pi_{f,z,y} \cdot \ell_{f,z,y}^t}_{\text{expected losses with instances (z,y)}}
$$
as the difference in expected loss between selecting right instances (z,y) and wrong instances (z,y'). 
\end{definition}

For example, $\alpha_{z,-}$ is the cross-instances cost of selecting the wrong MW instance (z,-) when the actual example has $y=+$. The cross-instances cost is bigger when weights learned by the wrong MW instance and the right MW instance are more disparate. In practice, since \model keeps track of weights $\pi_{f,z,y}$, $\alpha$ can be estimated and the estimation is updated at each round after the true label is revealed and individual classifier loss $\ell_{f,z,y}^t$ is realized.

\subsubsection{Regret Bound}

\begin{lemma}[Upper Bound]\label{lemma:upper_bound}
Let $f^*$ be the best expert in hindsight. The cumulative expected loss $\mathbb{E}[L] $ of \model can be bounded by:

\begin{align}\label{bound:regret_bound}
\mathbb{E}[L] \leq (1+\eta)L_{f^*} + 4\frac{\ln d}{\eta} + \alpha, 
\end{align}
where 
$
\displaystyle \alpha = \sum_{z \in \{A,B\},y \in \{+,-\}}\sum_t q_{z,y}^t\cdot\alpha_{z,y}^t.
$
\end{lemma}
This upper bound shows that the expected cumulative loss is bounded by the cumulative loss of the best classifier in hindsight plus the cumulative cross-instances cost $\alpha$.

In order to show the bound for differences in FPR across groups (i.e. for \eons), we also provide a lower bound on the expected cumulative loss of G-FORCE.

\begin{lemma}[Lower Bound]\label{lemma:lower_bound}
Let $f^*$ be the best expert in hindsight. 
Then, G-FORCE's expected cumulative loss is lower bounded by:
\begin{align}\label{bound:loss_bound}
    \mathbb{E}[L] \geq \gamma(\eta)\cdot L_{f^*} + \alpha.
\end{align}
where $\gamma(\eta)$ is defined as
$\displaystyle
\gamma(\eta) = \frac{\ln(1-\eta)}{\ln \left(1-\eta(1 + \eta)\right)}.
$
\end{lemma}

\newcommand{\Rvert}{\right\vert}
\newcommand{\Lvert}{\left\vert}

For the bound on fairness loss, each 
classifier $f \in \mathcal{F}$ satisfies some $\epsilon$-fairness with respect to data distribution $\mathbb{P}_{x,y,z}$, i.e., for $y \in \{+,-\}$;
\begin{equation}\label{eq:fair_iso}
\Lvert\mathbb{E}_{x,y,z}\left[\frac{L_{f,A,y}}{C_{A,y} }\right]- \mathbb{E}_{x,y,z}\left[\frac{L_{f,B,y}}{C_{B,y}}\right] \Rvert \leq \epsilon,
\end{equation}
where $C_{z,y}$ is the cardinality of group $z$ and label $y$.

\newcommand*\mystrut[1]{\vrule width0pt height0pt depth#1\relax}

\begin{theorem}[Fairness Bound]\label{theorem:fairness} Let $FPR_{f^*}(FNR_{f^*})$ be the classifier achieving lowest expected cumulative loss on subset $\{z,-\} (\{z,+\})$, $\forall z \in\{A,B\}$. For G-FORCE, and for $q_{A,-}\in [0,1]$ and $q_{B,-}\in [0,1]$, we have:
\begin{equation}\label{bound:fpr_bound}
\begin{aligned}
&\abs{ FPR_A - FPR_B } \\
&\leq \abs{ \left( 1+\eta - \gamma(\eta) \right)   FPR_{f^*} + \epsilon(1+\eta) + Q_{FPR}} \\
&\leq \abs{ G_{FPR}+ Q_{FPR}}
\end{aligned}
\end{equation}
\begin{equation}\label{bound:fnr_bound}
\begin{aligned} 
&\abs{ FNR_A - FNR_B }\\
&\leq \abs{ \left( 1+\eta - \gamma(\eta) \right)  FNR_{f^*} + \epsilon(1+\eta)+Q_{FNR}} \\
&\leq \abs{ G_{FNR}+ Q_{FNR}} .
\end{aligned}
\end{equation}

where $Q_{FPR}$ and $Q_{FNR}$ are functions of $\textbf{q} = [q_{A,-},q_{B,-},q_{A,+},q_{B,+}]^T$
\end{theorem}

\paragraph{Implication of the theoretical result}
The fairness bound shows the asymptotic result that after the optimization step converges, the absolute difference of \eo can be bounded by constants $G_{FPR}$ and $G_{FNR}$ . In appendix, we show that these constants depend on factors intrinsic to the problem: properties of the distribution and the fairness of the base classifiers ($\epsilon$,$FPR_{f^*}$,$FNR_{f^*}$). In appendix we also compare the theoretical bound of \eo with the achieved value of \eo in experiments to get a sense of tightness of the bound under different distributions.

\subsubsection{Optimal balance between regret and fairness}
\label{sec:optimal_Q}
In the appendix, we show that $Q_{FPR}$ and $Q_{FNR}$ can be set to zeros by solving the following set of equations:
\begin{align}\label{fpr_constraint_main}
\begin{bmatrix}
\displaystyle \frac{ \sum_{t}\alpha^t_{A,-}  }{p_A\cdot\mu_{A,-}\cdot T}   &   \displaystyle\frac{ -\sum_{t}\alpha^t_{B,-}}{p_B\cdot \mu_{B,-}\cdot T}
\end{bmatrix}
\begin{bmatrix}
    q_{A,-} \\
    q_{B,-}
\end{bmatrix}
= 0,
\end{align}

\begin{align}\label{fnr_constraint_main}
\begin{bmatrix}
\displaystyle \frac{-\sum_{t}\alpha^t_{A,+}  }{p_A\cdot\mu_{A,+}\cdot T}   &   \displaystyle\frac{\sum_{t}\alpha^t_{B,+}}{p_B\cdot \mu_{B,+} \cdot T}
\end{bmatrix}
\begin{bmatrix}
    q_{A,+} \\
    q_{B,+}
\end{bmatrix}
= 
0.
\end{align}

In addition, the upper bound for regret in Eq. (\ref{bound:regret_bound}) can also be tighten by adding the following constraint:
\begin{align}\label{regret_constraint_main}
\begin{bmatrix}
 \sum_{t}\alpha^t_{A,-} & \sum_{t}\alpha^t_{B,-} &  \sum_{t}\alpha^t_{A,+}& \sum_{t}\alpha^t_{B,+}
\end{bmatrix}
\mathbf{q} =  0,
\end{align}
where $\mathbf{q} = \begin{bmatrix} q_{A,-}  & q_{B,-} &  q_{A,+} & q_{B,+} & \end{bmatrix}^{T}$.

Given all these equations, constraints and inequalities we can define the following optimization step.
\paragraph{Optimization step} At each round, we are led to solve a linear system $\textbf{A}\textbf{q}=\textbf{b}$ where $\textbf{A}$ and $\textbf{b}$ are determined by the equations (\ref{fpr_constraint_main}), (\ref{fnr_constraint_main}) and (\ref{regret_constraint_main}) defined above. 
However, there are no prior guarantees of the existence of a solution, since matrix $\textbf{A}$ and vector $\textbf{b}$ are inherently defined by the given problem, i.e., the statistics of the data and the performance of the classifiers. 
Thus, we relax the condition to an optimization problem as:
\begin{align}\label{eq:optimization}
    \textbf{q}^* = \mbox{arg}\min_\textbf{q} || \boldsymbol{\lambda} ( \textbf{A}\textbf{q} - \textbf{b} )||_2^2 
\end{align}
where $\boldsymbol{\lambda}$ is a vector balancing the importance of equalized FPR, equalized FNR and regret that can be provided on a case-by-case basis for different applications. 
In our experiments, we solve (\ref{eq:optimization}) by using a Sequential Least Squares Programming method (SLSQP) and setting $\boldsymbol{\lambda} = \textbf{1}$.

In practice, \model can accommodate different use cases by setting different $\boldsymbol{\lambda}$ at each round. 
For example, during the early rounds, since the algorithm hasn't converged yet, we might want to set $\boldsymbol{\lambda}$ for equalized FPR and equalized FNR to be smaller to penalize the algorithm less for unfairness. Another scenario is a shifting distribution, and \model can be adaptive to the distribution with different $\boldsymbol{\lambda}$.

\section{Experiments and results}\label{experiments}
In this section we present G-FORCE's performance on real and synthetic datasets. 
\model keeps three statistics that are necessary to compute $Q_{FPR}$ and $Q_{FNR}$:
(i) the probability of a sample coming from group $z$, denoted by $p_z$, 
(ii) the \textit{base rates of outcomes}, denoted by $\mu_{z,y}$, and 
(iii) the cross-instance costs $\alpha$, which is estimated as differences of expected loss between using a right instance and a wrong instance. All three statistics above are estimated with Bayesian and Dirichlet Prior. We use $\eta = 0.35$ in experiments.
\subsection{Case study: Synthetic Data sets}
It is important to test what can be achieved for both algorithms under extreme scenarios. We create a synthetic data framework that allows us to control the distributions and experts with certain properties. 

\begin{figure}[h]
  \centering
 \vspace*{-0.15in}
  \subfloat[Imbalanced Setting.\label{fig:tree_plot(a)}]{\includegraphics[width=0.4\linewidth]{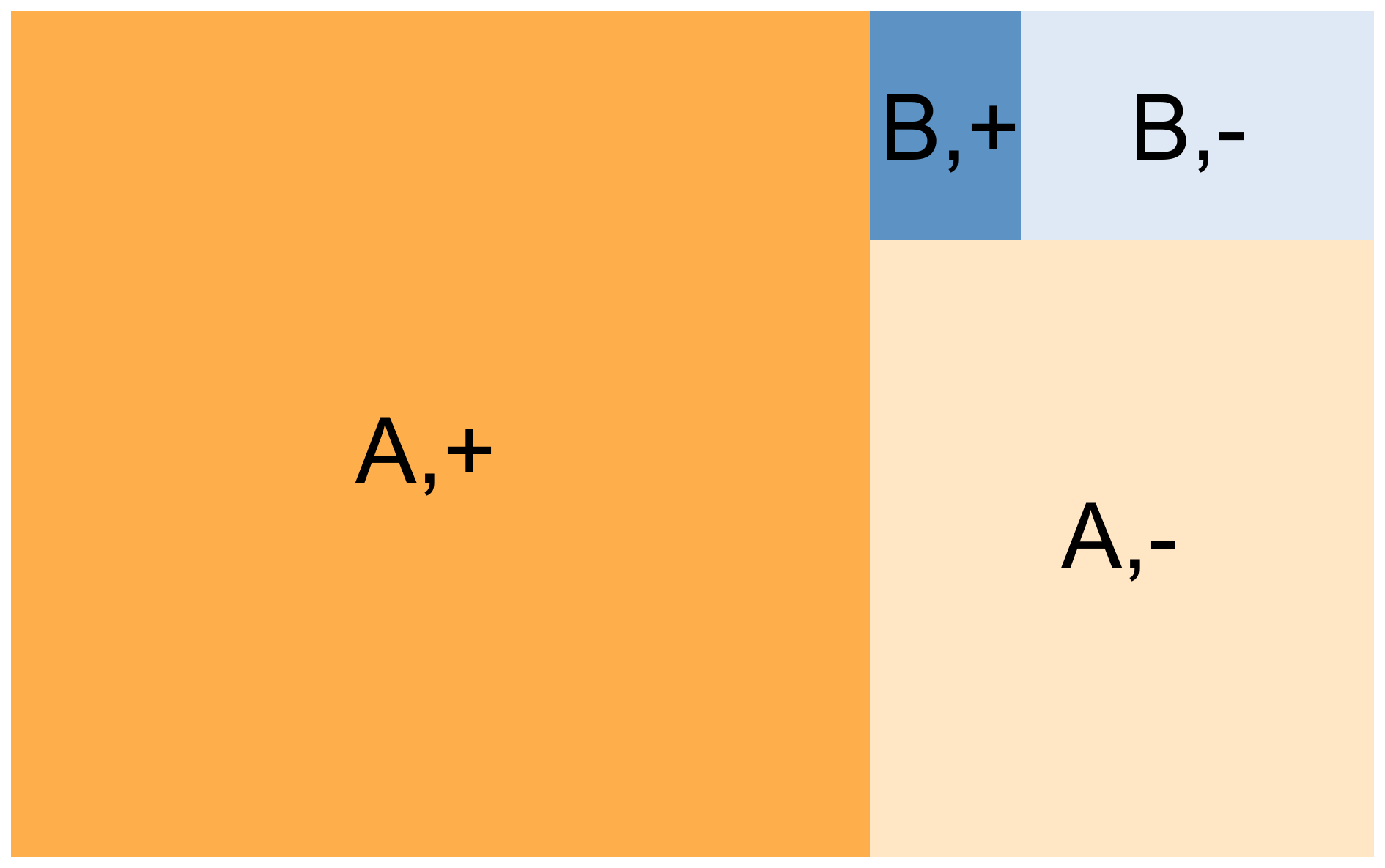}}
\hspace{0.5cm}
  \subfloat[Balanced Setting.
  \label{fig:tree_plot(b)}]{\includegraphics[width=0.4\linewidth]{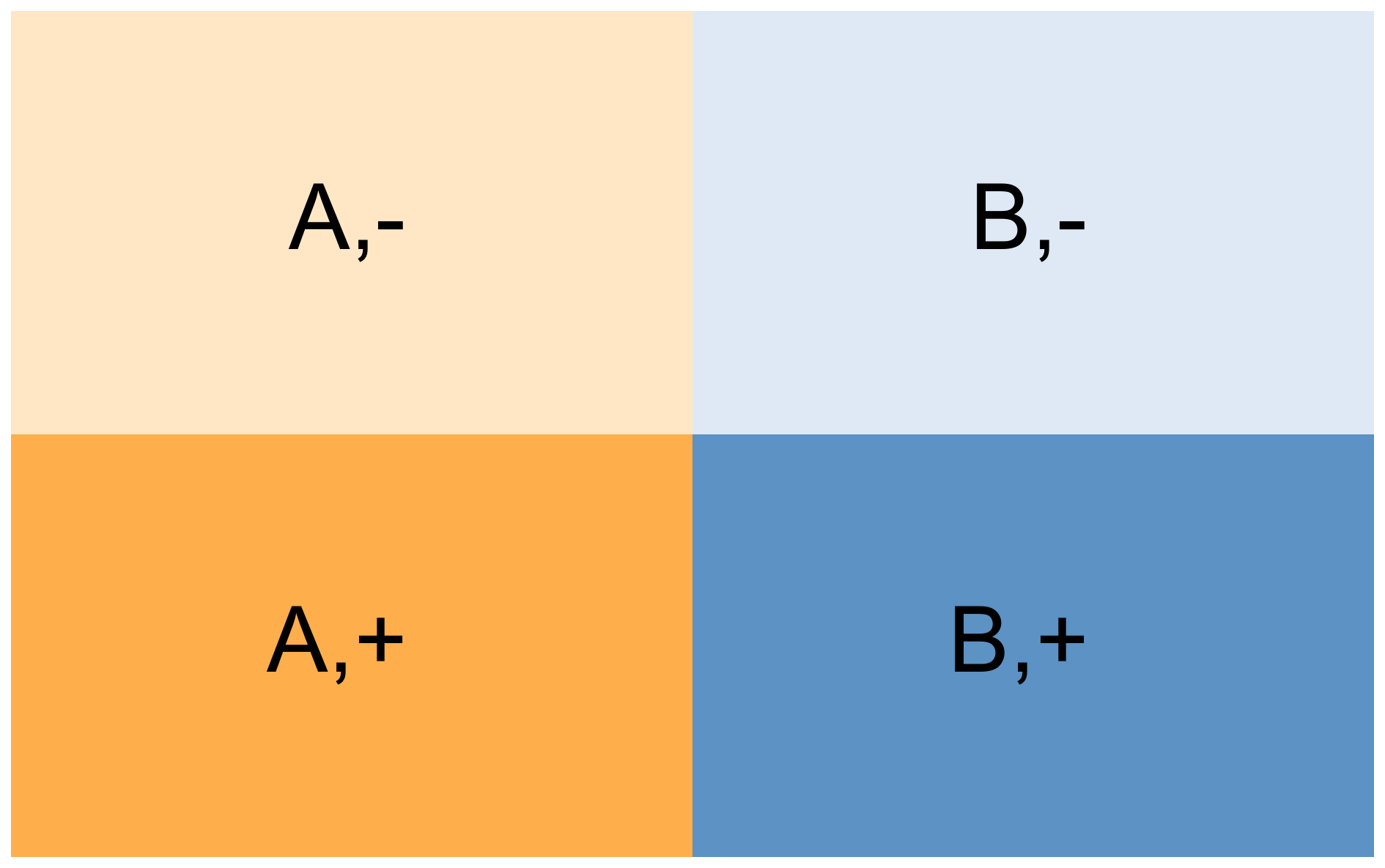}}
  \caption{The size of each color block is proportional to the number of examples in that group-label subset. Imbalanced setting is created with $p_A=0.9,\mu_{A,+}=0.7,\mu_{B,+}=0.3$ and balanced setting is created with $p_A=0.5,\mu_{A,+}=0.5,\mu_{B,+}=0.5$.}
  \label{fig:tree_plot}
\end{figure}

The balance between protected groups and labels is controlled by setting parameters $p_A,\mu_{A,+},\mu_{B,+}$. We visualize two such settings in Figure \ref{fig:tree_plot}. For each dataset, we repeat the experiments 100 times, each with 10000 samples from a specific distribution setting. 

It is also important to test the efficacy of our approach when experts have disparate performances or are extremely biased towards different groups. For binary classification with two protected groups, we create four ``biased'' classifiers/experts, where each is perfect ($100\%$ accurate) for one of the group-label subsets ($\{A,+\},\{A,-\},\{B,+\},\{B,-\}$), and random ($50\%$ accurate) for the other three. Thus for each group-label subset, there is at least one perfect expert/classifier.

\begin{figure}[ht]
  \vspace*{-0.15in}
    \centering
  \subfloat[Imbalanced Setting.\label{fig:bar_plot(a)}]{\includegraphics[width=0.9\linewidth]{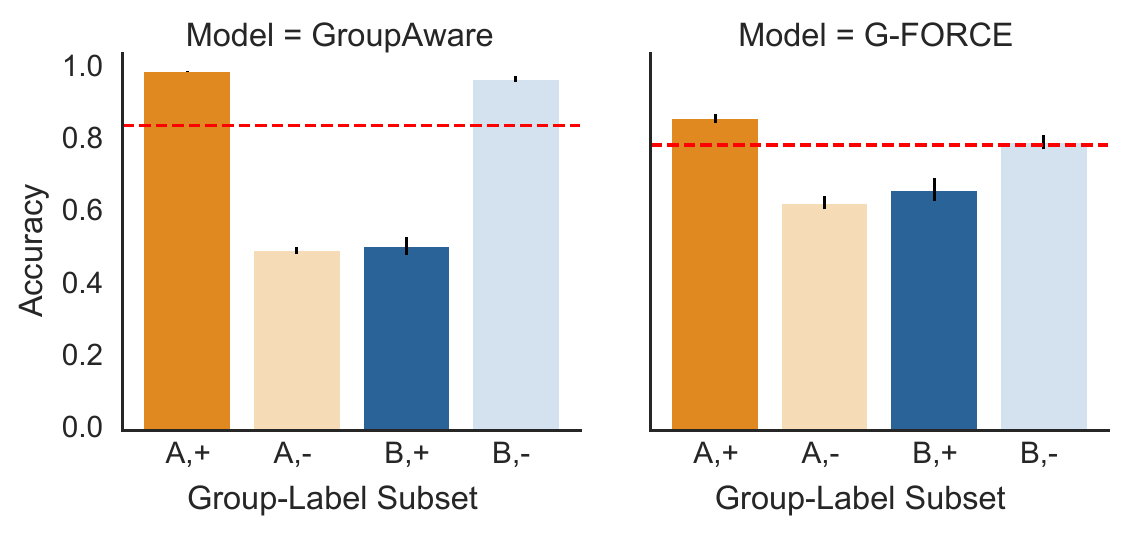}}\\
  \subfloat[Balanced Setting. \label{fig:bar_plot(b)}]{\includegraphics[width=0.9\linewidth]{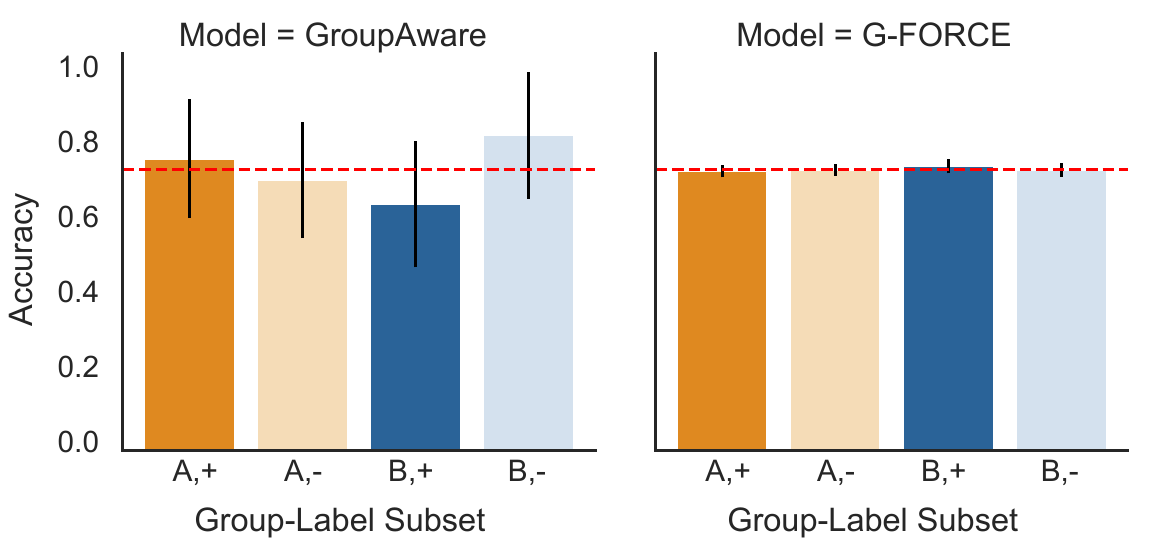}}
    \caption{The achieved accuracy on group-label subsets for imbalanced setting ($p_A=0.9,\mu_{A,+}=0.7,\mu_{B,+}=0.3$) and balanced Setting ($p_A=0.5,\mu_{A,+}=0.5,\mu_{B,+}=0.5$). Left: GroupAware. Right: G-FORCE. The vertical black line denotes the standard deviation. The red dashed line is the overall accuracy. }
\end{figure}

For imbalanced setting in Figure~\ref{fig:tree_plot(a)}, the results in Figure~\ref{fig:bar_plot(a)} shows that for \groupaware algorithm,  the larger subsets $\{A,+\}$ and $\{B,-\}$ have nearly $100\%$ accuracy while $\{A,-\}$ and $\{B,+\}$ have around $50\%$ accuracy. The \groupaware algorithm, which runs only one MW instance per sensitive group, promotes selecting the perfect classifier for the larger group-label subset within each protected group. This leads to high error rates on the remaining subsets since their associated perfect classifiers are unlikely to be picked. 

Even for a perfectly balanced setting (Figure \ref{fig:tree_plot(b)}), \model achieves a more balanced accuracy in each subset and a more stable behavior compared to \groupaware as in Figure~\ref{fig:bar_plot(b)}. Since the label distribution is balanced, $\{A,-\}$ and $\{A,+\}$ have the same accuracy when classifying an example from group A.
\groupaware arbitrarily chooses between perfect classifier for $\{A,+\}$ or $\{A,-\}$ when classifying examples from group A, which leads to large deviations when considering errors on each more fine-grained subset (same analogy for group B).
On the contrary, in both settings, \model is able to track the performance of the \eo on each group-label subset and compensate their differences in terms of accuracy. The side effect, as expected, is a slight decrease in accuracy. 
Due to the lack of space, we leave more experiment results in appendix. In appendix we also plot Pareto Curve to characterize the trade-off that can be achieved in G-FORCE.

\subsection{Case study: Real Data sets}

\begin{table*}[t]
\centering
\resizebox{0.95\textwidth}{!}{
\begin{tabular}{c|c|c|c|c|c|c|c|c|c}
\toprule
           & \multicolumn{3}{c|}{\texttt{Adult}}  & \multicolumn{3}{c|}{\texttt{Compas}} & \multicolumn{3}{c}{\texttt{German}} 
           \\ \cline{2-10}
           &  FPR & FNR & Regret & FPR & FNR& Regret & FPR           & FNR                                        & Regret  \\
        \cline{2-10} 
\groupaware & 0.05$\pm$ 0.01      & 0.17$\pm$ 0.02 & \textbf{0.00}$\pm$ 0.00 & 0.20$\pm$ 0.04                          & 0.27$\pm$ 0.04                           & \textbf{0.01}$\pm$ 0.00 & 0.40$\pm$ 0.13                          & 0.21$\pm$ 0.08                           & 0.01$\pm$ 0.01 \\
\cline{1-10} 
\model      & \textbf{0.04}$\pm$ 0.01 & \textbf{0.08}$\pm$ 0.01 & 0.01$\pm$ 0.00                        & \textbf{0.18}$\pm$ 0.03 & \textbf{0.25}$\pm$ 0.04 & 0.01$\pm$ 0.01                           & \textbf{0.32}$\pm$ 0.15 & \textbf{0.18}$\pm$ 0.01 & 0.01$\pm$ 0.01    \\
\bottomrule
\end{tabular}
}
\caption{Equalized FPR, equalized FNR and regret on real datasets. Lower numbers are better.}
\label{tab:real_data}
\end{table*}
\paragraph{Datasets} We consider the \texttt{Adult}, \texttt{German Credit} and \texttt{COMPAS} datasets, all of which are commonly used by the fairness community. \texttt{Adult} consists of individuals' annual income measurements based on different factors. In the \texttt{German} dataset, people applying for credit from a bank are classified as ``good'' or ``bad'' credit based on their attributes. \texttt{COMPAS} provides a likelihood of recidivism based on a criminal defendant’s history.

\paragraph{Creating Black-box Experts} The set of classifiers $\mathcal{F}$ that form the black-box experts are: Logistic Regression (LR), Linear SVM (L SVM), RBF SVM, Decision Tree (DT) and Multi-Layer Perceptron (MLP). These classifiers are trained using 70\% of the data set. The remaining 30\% of the dataset is set aside to simulate the online arrival of individuals. 
We compare our \model algorithm with the \groupaware in terms of regret and fairness. We repeated the experiments 1000 times for \texttt{German} and \texttt{COMPAS}, as well as 10 times for \texttt{Adult}, by randomizing the arrival sequence of individuals.  

\begin{figure}[h]
\centering
\vspace*{-0.15in}

 \subfloat[Equalized FPR]{\includegraphics[width=0.45\linewidth]{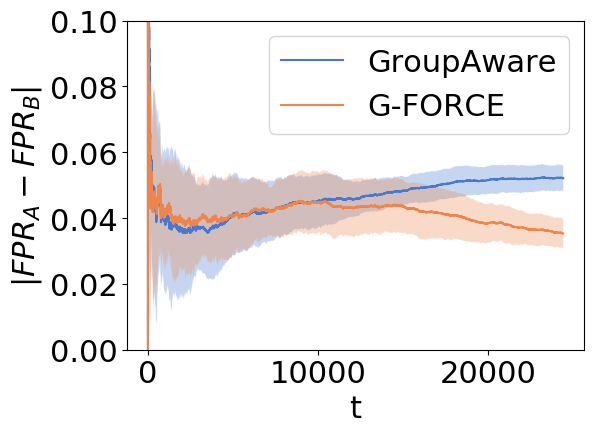}}
  \subfloat[Equalized FNR]{\includegraphics[width=0.45\linewidth]{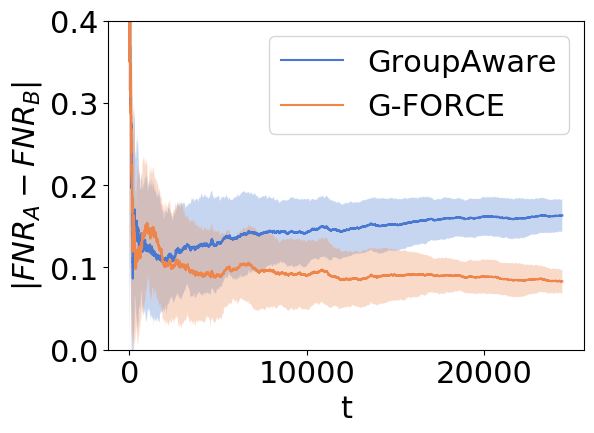}}

\caption{\model shows a clear improvement over GroupAware on both equalized FPR (left) and equalized FNR (right).}
\label{fig:adult}
\end{figure}

Results in Table-\ref{tab:real_data} show a general improvement in fairness over the \groupaware algorithm, both in terms of equalized FPR and FNR, along with a small increase in regret. For \texttt{Adult} data set, we plot the performance of the algorithm over time (Figure \ref{fig:adult}). Although \texttt{German} and \texttt{COMPAS} have fewer examples, and thus the standard deviation is higher to make a conclusion, there is still a slight improvement over fairness with slight increase in regret. 

  We also report the error rates and associated $\epsilon$-fairness of each classifier in the appendix. The base classifiers expose similar and more mild behaviors (compared with in real datasets) which makes the task of the algorithm easier, and thus the results are less significant compared to the real dataset. We would like to highlight again that the goal of the paper is not to propose an algorithm always better in experiments than \groupaware (especially since we don't put assumption on $\epsilon$-fairness), but is to have an algorithm with \textbf{guaranteed behaviors} under \textbf{any stochastic scenarios}.

\section{Discussion}\label{discussion}

In this paper, we introduce G-FORCE, a randomized MW-based algorithm achieving approximate \eons. Our algorithm gives a provable bound for the number of false positives and negatives obtained in an online stochastic setting, which could be potentially useful beyond the intended application of achieving fairness. We believe \model can be applied to a wide range of applications as it could work alongside with human decision makers and correct potential biases. A user could choose a $\lambda$ to set a desirable trade-off between fairness and accuracy. We are also deploying the algorithm to a real world application. 

\paragraph{Future work} Future research could take on a more realistic case in which feedback is delayed for some number of rounds.
For example, during the college admissions process, the performance of a student is generally evaluated at the end of each term, while colleges typically offer admission decisions in mid-year. Similarly, when an individual applies for a loan, the bank often needs to wait for some time to know whether the applicant will default or not.

\section*{Acknowledgements}
Dr. Cuesta-Infante is funded by the Spanish Government research fundings RTI2018-098743-B-I00 (MICINN/FEDER) and Y2018/EMT-5062 (Comunidad de Madrid)
\bibliographystyle{named}
\bibliography{ijcai21}

\clearpage
\onecolumn
\section{Appendixes}
\section*{Additional Experiment Results}
\subsection{Synthetic Data Set}
\subsubsection{Additional Experiments: Pareto Curve on Synthetic Dataset}
\begin{figure}[!htbp]
    \centering
    \includegraphics[scale=0.6]{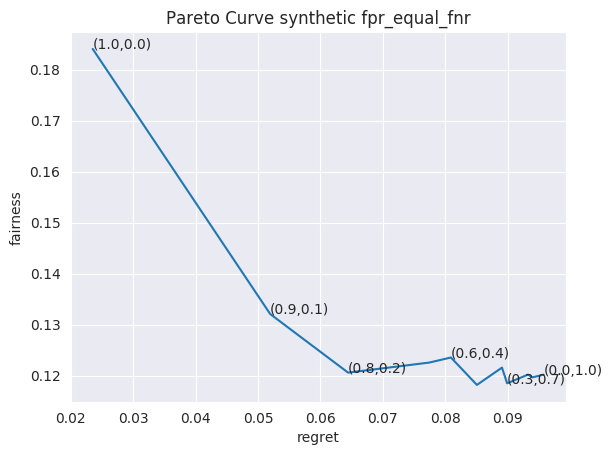}
    \caption{Pareto Curve for the synthetic dataset with imbalanced setting. x-axis is the regret and y-axis is the average value of Equalized FPR and Equalized FNR. The pair indicates ($\boldsymbol{\lambda}_{regret},\boldsymbol{\lambda}_{Fairness}$) where $\boldsymbol{\lambda}_{Fairness} = \boldsymbol{\lambda}_{FPR} = \boldsymbol{\lambda}_{FNR}$.}
    \label{fig:pareto_curve}
\end{figure}

To clearly illustrate the trade-off that can be achieved in the optimization step, we plot the pareto front by varying $\boldsymbol{\lambda}$ defined in the optimization step \ref{eq:optimization}. The Pareto curve is in \ref{fig:pareto_curve}.

\subsubsection{Additional Experiments: different distribution on synthetic dataset}
We summarize the experiments for \texttt{Synthetic} data in Tables \ref{tab:synth_fpr} and \ref{tab:synth_fnr}, where we fix $p_A = 0.9, p_B = 0.1, \mu_{A,+} = 0.7$ and varies $\mu_{B,+}$. 

\begin{table}[!htbp]
\centering
\begin{tabular}{lrrr}
  \toprule
  $\mu_{B,+}$ &    MW &  \groupaware &  \model \\
  \midrule
0.1 & $0.016\pm 0.013$ & $0.494\pm 0.009$ & $\textbf{0.305}\pm 0.020$ \\
0.3 & $0.018\pm 0.013$ & $0.487\pm 0.012$ & $\textbf{0.182}\pm 0.014$ \\
0.4 & $0.026\pm 0.011$ & $0.475\pm 0.019$ & $\textbf{0.148}\pm 0.032$ \\
0.5 & $0.024\pm 0.018$ & $0.283\pm 0.162$ & $\textbf{0.110}\pm 0.030$ \\
0.6 & $0.011\pm 0.008$ & $\textbf{0.019}\pm 0.022$ & $0.032\pm 0.017$ \\
\bottomrule
\end{tabular}
\caption{Equalized FPR by fixing $p_A = 0.9,p_B = 0.1,\mu_{A,+}=0.7$}
\label{tab:synth_fpr}
\end{table}

\begin{table}[!htbp]
\centering
\begin{tabular}{lrrr}
  \toprule
  $\mu_{B,+}$ &    MW &  \groupaware &  \model \\
  \midrule
0.1 & $0.473\pm 0.060$ & $0.509\pm 0.055$ & \textbf{0.304}$\pm$ 0.028 \\
0.3 & $0.490\pm 0.031$ & $0.486\pm 0.022$ & \textbf{0.194}$\pm$ 0.018 \\
0.4 & $0.508\pm 0.018$ & $0.488\pm 0.019$ & \textbf{0.146}$\pm$ 0.018 \\
0.5 & $0.488\pm 0.013$ & $0.296\pm 0.162$ & \textbf{0.111} $\pm$ 0.030 \\
0.6 & $0.495\pm 0.020$ & $\textbf{0.022}\pm 0.019$ & $0.046\pm 0.010$ \\
\bottomrule
\end{tabular}
\caption{Equalized FNR by fixing $p_A = 0.9,p_B = 0.1,\mu_{A,+}=0.7$}
\label{tab:synth_fnr}
\end{table}

\subsection{Real Data Sets}
\subsubsection{Summary statistics of real data sets}

\begin{table}[H]
\centering
\begin{tabular}{ccclll}
\hline
                          & \# of rounds             & $p_A$                       & $\mu_{A,+}$ & $\mu_{B,+}$  \\ \hline
\texttt{Adult}                      & 24421                    & 0.851                     & 0.26                  & 0.16                                                                    \\
\texttt{German Credit }             & 300                      & 0.853                     & 0.73                  & 0.50                                                                     \\
\texttt{COMPAS} & 1584 & 0.398 & 0.54                  & 0.39                                                                   \\ \hline
\end{tabular}
\end{table}

\begin{table}[H]
\centering
\begin{tabular}{@{}cc|ccccc|ccc@{}}
\multicolumn{2}{c}{} & \multicolumn{5}{c}{Individual Experts} & \multicolumn{2}{c}{Combined Experts}\\
\toprule
            &           &  L SVM & RBF SVM & DT & MLP & LR & Group-Aware & \model  \\ \midrule
       & FPR &  0.022 &         0.046 & 0.043 & 0.047 & 0.047   & 0.052 &   \textbf{0.035} \\
\texttt{Adult} & FNR &  0.026& 0.199& 0.200& 0.214& 0.214                 & 0.163 &  \textbf{0.083} \\
     & EER &  0.058& 0.062& 0.062& 0.061& 0.061               & 0.074 &  \textbf{0.069} \\
  \hline
     & FPR &  0.00&         0.371& 0.471& 0.421& 0.050 & 0.373 &   \textbf{0.329} \\
 \texttt{German} & FNR & 0.000 &         0.320 & 0.770 & 0.680 & 0.650    & 0.207 &  \textbf{0.181} \\
     & EER &  0.090 &         0.090 & 0.208 & 0.210 & 0.280 &  \textbf{0.093} &  0.098\\
\hline
    & FPR        & 0.190     & 0.150   & 0.160         & 0.158      & 0.240       & 0.191 &  \textbf{0.184} \\
\texttt{COMPAS} & FNR        & 0.256      & 0.240   & 0.260         & 0.240      & 0.340      & 0.264 &  \textbf{0.249} \\
     & EER        & 0.019      & 0.010   & 0.010         & 0.010      & 0.010      & \textbf{0.016} &  0.019 \\

\bottomrule
\end{tabular}
\caption{$\epsilon$-Fairness of base experts, \groupaware  and \model. }
\label{base fairness}
\end{table}




\subsection{Proofs}
We define the cumulative loss of classifier $f$ on group $z$ as 
$L_{f,z} = \sum_{t=1}^T \ell_{f,z}^t$. 
The cumulative false positive of $f$ on group $z$ is the cumulative loss made on the negative examples; and its expression is 
$L_{f,z,-} = \sum_{t=1}^T \ell_{f,z}^t\mathds{1}\{y = -\}$. 
Similarly, we defined the expected loss on group $z$ as 
$\mathbb{E}[L_z] = \sum_{t=1}^T\sum_{f \in \mathcal{F}}\pi_f^t\ell_{f,z}^t$ 
and the expected false positive on group $z$ is 
$\mathbb{E}[L_{z,-}] = \sum_{t=1}^T\sum_{f \in \mathcal{F}}\pi_f^t\ell_{f,z}^t\mathds{1}\{y=-\}$. 
\subsubsection{Proof of Lemma 1}
Let $\Phi_{z,+}^{t} =  \sum_{f \in \mathcal{F}}w_{f,z,+}^{t}$. We start computing the expected loss on group ${z,+}$:
\begin{align}\label{eq:org}
    \mathbb{E}[ \ell^t_{z,+} ] &= \sum_{f \in \mathcal{F}} \pi_{f,z}^t \cdot \ell_{f,z}^t\mathds{1}\{y = +\}  \nonumber \\ 
    &=  \sum_{f \in \mathcal{F}}  \left(q_{z,+}^t \cdot \frac{\displaystyle w_{f,z,+}}{ \displaystyle \sum_{f \in \mathcal{F}} w_{f,z,+}} + q_{z,-}^t \cdot \frac{\displaystyle w_{f,z,-}}{ \displaystyle \sum_{f \in \mathcal{F}} w_{f,z,-}}\right) \cdot \ell_{f,z}^t\mathds{1}\{y = +\}  \nonumber\\
    &=  q_{z,+}^t \cdot \sum_{f \in \mathcal{F}} \frac{\displaystyle w_{f,z,+}^t}{ \Phi_{z,+}^{t} } \cdot \ell_{f,z}^t\mathds{1}\{y = +\} + q_{z,-}^t \cdot \sum_{f \in \mathcal{F}} \frac{\displaystyle w_{f,z,-}^t}{ \Phi_{z,-}^{t} } \cdot \ell_{f,z}^t\mathds{1}\{y = +\} \nonumber\\
    &=  q_{z,+}^t \cdot \sum_{f \in \mathcal{F}} \frac{\displaystyle w_{f,z,+}^t}{ \Phi_{z,+}^{t} } \cdot \ell_{f,z,+}^t  + q_{z,-}^t \cdot \sum_{f \in \mathcal{F}} \frac{\displaystyle w_{f,z,-}^t}{ \Phi_{z,-}^{t} } \cdot \ell_{f,z,+}^t
\end{align}
The overall expected loss is composed by two terms: the former, which is the expected loss on group $z,+$ when their associated weights $w_{f,z,+}$ are selected, and the later, when the wrong weights $w_{f,z,-}$ are selected. Both terms are weighted by their respective estimated rates $q_{z,+}^t$ and $q_{z,-}^t$.

Then, we have the following inequality:

\begin{align*}
    \Phi_{z,+}^{t+1} &=  \sum_{f \in \mathcal{F}} w_{f,z,+}^{t+1} \nonumber \\
    &= \sum_{f \in \mathcal{F}} w_{f,z,+}^t(1-\eta)^{\ell_{f,z}^t\mathds{1}\{y = +\}} \nonumber \\
    & \leq \sum_{f \in \mathcal{F}} w_{f,z,+}^t(1-\eta \ell_{f,z}^t\mathds{1}\{y = +\}) \nonumber \\
    &= \sum_{f \in \mathcal{F}}  w_{f,z,+}^t - \eta \sum_{f \in \mathcal{F}}w_{f,z,+}^t  \ell_{f,z,+}^t \nonumber\\
    &= \Phi_{z,+}^{t}(1  - \eta \sum_{f \in \mathcal{F}} \frac{w_{f,z,+}^t}{\Phi_{z,+}^{t}}\cdot \ell_{f,z,+}^t)
\end{align*}
Thus by the recursive function, we have
\begin{align}\label{eq:upper1}
 \Phi_{z,+}^{T+1} &\leq  \Phi_{z,+}^{1}\prod_{t=1}^T (1  - \eta \sum_{f \in \mathcal{F}} \frac{w_{f,z,+}^t}{\Phi_{z,+}^{t}}\cdot \ell_{f,z,+}^t) \nonumber \\ 
&= d \prod_{t=1}^T (1  - \eta \sum_{f \in \mathcal{F}} \frac{w_{f,z,+}^t}{\Phi_{z,+}^{t}}\cdot \ell_{f,z,+}^t).
\end{align}

Following the updating rule of the MW algorithm, we have
\begin{align}\label{eq:lower1}
w_{f,z,+}^{T+1} &=  w_{f,z,+}^1 (1 - \eta)^{\sum_{t=1}^T \ell_{f,z}^t \cdot \mathds{1}\{y = +\}} \nonumber\\
& = (1 - \eta)^{\sum_{t=1}^T \ell_{f,z,+}^t}
\end{align}
where $w_{f,z,+}^t = 1$, as all the weights are initialized.

Using \ref{eq:upper1} and \ref{eq:lower1},
\begin{equation}\label{bounds1}
 w^{T+1}_{f,z,+}= (1 - \eta)^{\sum_{t=1}^T \ell_{f,z,+}^t } \leq \Phi_{z,+}^{T+1}\leq d \prod_{t=1}^T (1  - \eta \sum_{f \in \mathcal{F}} \frac{w_{f,z,+}^t}{\Phi_{z,+}^{t}}\cdot \ell_{f,z,+}^t)
\end{equation}
and taking the logarithm of both sides, we have
\begin{align}
\ln(1-\eta)L_{f,z,+} &\leq \ln d + \sum_{t=1}^T \ln(1  - \eta \sum_{f \in \mathcal{F}} \frac{w_{f,z,+}^t}{\Phi_{z,+}^{t}}\cdot \ell_{f,z,+}^t)
\nonumber\\
\ln(1-\eta)L_{f,z,+} &\leq \ln d -\eta \sum_{f \in \mathcal{F}} \frac{w_{f,z,+}^t}{\Phi_{z,+}^{t}}\cdot \ell_{f,z,+}^t \label{eq:ineq}
\\
\sum_{t=1}^T \sum_{f \in \mathcal{F}} \frac{w_{f,z,+}^t}{\Phi_{z,+}^{t}}\cdot \ell_{f,z,+}^t &\leq (1+\eta)L_{f,z,+}+\frac{\ln d}{\eta}\label{eq:ineq2}
\end{align}
Equation \ref{eq:ineq} follows because if $\eta < 1/2$, we can use the inequality $\ln(1-\eta)<-\eta$. This is intuitive as if we always choose weights for positive examples, it reduces to the same bound as in the original MW algorithm.

We now assume that the expected error on group ${z,+}$ when wrong weights $w_{f,z,-}$ are selected is bounded as:
\begin{align}\label{eq:alpha}
\sum_{f \in \mathcal{F}} \frac{\displaystyle w_{f,z,-}^t}{ \Phi_{z,-}^{t} } \cdot \ell_{f,z,+}^t \leq \sum_{f \in \mathcal{F}} \frac{\displaystyle w_{f,z,+}^t}{ \Phi_{z,+}^{t} } \cdot \ell_{f,z,+}^t + \alpha^{t}_{z,-}
\end{align}
where $\alpha^{t}_{z,-}<1$ is the difference of loss in expectation made when using the incorrect weights of the MW algorithm on group ${z,+}$ (Cross-Instance Cost). Then
\begin{align}
\mathbb{E}[L_{z,+}] & = \sum_{t=1}^{T} \left( q_{z,+}^t \cdot \sum_{f \in \mathcal{F}} \frac{\displaystyle w_{f,z,+}^t}{ \Phi_{z,+}^{t} } \cdot \ell_{f,z,+}^t  + q_{z,-}^t \cdot \sum_{f \in \mathcal{F}} \frac{\displaystyle w_{f,z,-}^t}{ \Phi_{z,-}^{t} } \cdot \ell_{f,z,+}^t \right)\\
\mathbb{E}[L_{z,+}] & \leq   \sum_{t=1}^{T} \sum_{f \in \mathcal{F}} \frac{\displaystyle w_{f,z,+}^t}{ \Phi_{z,+}^{t} } \cdot \ell_{f,z,+}^t  +  \sum_{t} q_{z,-}^t \cdot\alpha^t_{z,-} 
\end{align}
where using \ref{eq:ineq2}, we finally obtain:
\begin{align}
    \mathbb{E}[L_{z,+}] & \leq (1+\eta)L_{f,z,+} + \frac{\ln d}{\eta} +  \sum_{t} q_{z,-}^t\cdot \alpha^t_{z,-}
\end{align}
Similarly,
\begin{align}
    \mathbb{E}[L_{z,-}] & \leq  (1+\eta)L_{f,z,-} + \frac{\ln d}{\eta} + \sum_{t} q_{z,+}^t \cdot \alpha^t_{z,+}.
\end{align}
The expected total errors on group $z$ is, adding the two equations above:
\[ \mathbb{E}[L_{z}] \leq (1+\eta)L_{f,z}+2\frac{\ln d}{\eta} + ( \sum_{t} q_{z,-}^t \cdot\alpha^t_{z,-} + \sum_{t} q_{z,+}^t \cdot \alpha^t_{z,+})\].

In the same way, the expected total errors (considering $z = {A,B}$) is:
\begin{align}\label{eq:loss_bound}
\mathbb{E}[L] \leq (1+\eta)L_{f} + 4\frac{\ln d}{\eta} + \alpha 
\end{align}

where all the Cross-Instance Costs are condensed in:
$$
\alpha = \sum_{z \in \{A,B\},y \in \{-,+\}}q_{z,y}\sum_t\alpha_{z,y}^t.
$$

\subsubsection{Proof of Lemma 2}
Using the same process as for the upper bound, we have:

\begin{align*}
    \Phi_{z,+}^{t+1} &=  \sum_{f \in \mathcal{F}} w_{f,z,+}^{t+1}\\
    &= \sum_{f \in \mathcal{F}} w_{f,z,+}^t(1-\eta)^{\ell_{f,z}^t \mathds{1}\{y=+\}} \\
    & \geq \sum_{f \in \mathcal{F}} w_{f,z,+}^t(1-\eta(1 + \eta) \ell_{f,z}^t\mathds{1}\{y = +\}\\
    &= \sum_{f \in \mathcal{F}}  w_{f,z,+}^t - \eta(1 + \eta) \sum_{f \in \mathcal{F}}w_{f,z,+}^t  \ell_{f,z,+}^t\\
    &= \Phi_{z,+}^{t}(1  -\eta(1 + \eta)\sum_{f \in \mathcal{F}} \frac{ w_{f,z,+}^t}{\Phi^t_{z,+}}\cdot \ell_{f,z,+}^t).
\end{align*}
Thus, by the recursive function, we have
\begin{align*}
 \Phi_{z,+}^{T+1} &\geq  \Phi_{z,+}^{1}\prod_{t=1}^T (1  - \eta(1 + \eta)  \sum_{f \in \mathcal{F}} \frac{ w_{f,z,+}^t}{\phi^t_{z,+}}\cdot \ell_{f,z,+}^t) \\ 
&= d \prod_{t=1}^T (1  - \eta(1 + \eta)\sum_{f \in \mathcal{F}} \frac{ w_{f,z,+}^t}{\Phi^t_{z,+}}\cdot \ell_{f,z,+}^t)\\
\end{align*}.

Let $f^*$ be the best expert in hindsight in terms of achieving lowest false positives, we have
\begin{align*}
    \Phi_{z,+}^t &= \sum_{f \in \mathcal{F}}w_{f,z,+}^t  \\
    &\leq d\cdot \max_{f \in \mathcal{F}}w_{f,z,+}^t \\
    &= d\cdot w_{f^*,z,+}^t \\
    &= d \cdot \max_{f \in \mathcal{F}}(1 - \eta)^{\sum_{t=1}^T \ell_{f,z}^t \cdot \mathds{1}\{y = +\}} \\
    &= d \cdot (1 - \eta)^{\sum_{t=1}^T \ell_{f^*,z}^t \cdot \mathds{1}\{y = +\}}. \\
\end{align*}
Therefore we have:
\[d \cdot (1 - \eta)^{\sum_{t=1}^T \ell_{f^*,z}^t \cdot \mathds{1}\{y = 1\}} \geq \Phi_{z,+}^t \geq d\cdot \prod_{t=1}^T[1  - \eta(1+\eta) \sum_{f \in \mathcal{F}} \frac{ w_{f,z,+}^t}{\Phi^t_{z,+}}\cdot \ell_{f,z,+}^t]
\]
 Taking the log of both sides:
\begin{align}
    \ln(1-\eta)L_{f^*,z,+} &\geq \sum_{t=1}^T \ln\left(1  - \eta(1+\eta)  \sum_{f \in \mathcal{F}} \frac{ w_{f,z,+}^t}{\Phi^t_{z,+}}\cdot \ell_{f,z,+}^t\right) \\
    \ln(1-\eta)L_{f^*,z,+} &\geq  \ln \left(1  - \eta(1+\eta)\right)  \sum_{t=1}^T \sum_{f \in \mathcal{F}} \frac{ w_{f,z,+}^t}{\Phi^t_{z,+}}\cdot \ell_{f,z,+}^t\\
    \sum_{t=1}^T \sum_{f \in \mathcal{F}} \frac{ w_{f,z,+}^t}{\Phi^t_{z,+}}\cdot \ell_{f,z,+}^t &\geq \gamma(\eta)L_{f^*,z,+}
\end{align}
where $\gamma(\eta)$ is defined as:
$$
\gamma(\eta) = \frac{\ln(1-\eta)}{\ln \left(1-\eta(1 + \eta)\right)}
$$
using that $ln(1 -\eta(1+\eta)x) \geq ln(1 - \eta(1+\eta))x$ for all $x \in [0,1]$ and $\eta \in (0,\eta^{max})$, where $\eta^{max} = \frac{-1+\sqrt{5}}{2}$ which does not restrict the range of $\eta \in (0,0.5)$.

Thus, using \ref{eq:alpha}, we have:
\begin{align*}
    \mathbb{E}[L_{z,+}] &= \sum_{t=1}^T (q_{z,+}^t \sum_{f \in \mathcal{F}} \frac{ w_{f,z,+}^t}{\Phi^t_{z,+}}\cdot \ell_{f,z,+}^t+q_{z,-}^t \sum_{f \in \mathcal{F}} \frac{ w_{f,z,-}^t}{\Phi^t_{z,-}}\cdot \ell_{f,z,+}^t \\
    &\geq \gamma(\eta)L_{f^*,z,+} + \sum_{t} q_{z,-}^t \cdot\alpha^t_{z,-}.
\end{align*}
and,
\begin{align*}
    \mathbb{E}[L_{z}]
    &\geq \gamma(\eta)L_{f^*,z} + (\sum_{t}q_{z,-}^t \cdot \alpha^t_{z,-} + \sum_{t}q_{z,+}^t \cdot \alpha^t_{z,+}).
\end{align*}
Finally, the total expected error is lower bounded by:
\begin{align}
        \mathbb{E}[L]
    &\geq \gamma(\eta)L_{f^*} + \alpha.
\end{align}
\subsubsection{Fairness Bound}
\paragraph{Proof} 
We assume group A arrives with probability p, group B arrives with probability 1-p, that is, $\mathbb{P}(z=A) = p$. 
The expected mean label of group A is defined as $\mu_{A,+} = \mathbb{P}(y=+|z=A)$ and mean label of group B is defined as $\mu_{B,+} = \mathbb{P}(y=+|z=B)$.
Each individual classifier is $\epsilon$-fair, thus:
\begin{align}\label{eq:epfair}
\Lvert\mathbb{E}_{x,y,z}\left[\frac{L_{f,A,-}}{\sum_{t=1}^{T}\mathds{1}
\{y=-\}\mathds{1}\{z=A\} }\right]-\mathbb{E}_{x,y,z}\left[\frac{L_{f,B,-}}{\sum_{t=1}^{T}\mathds{1}
\{y=-\}\mathds{1}\{z=B\}}\right]\Rvert\leq \epsilon,  ~\forall f
\end{align}
which represents the cardinality of the selected subset of samples.

The absolute difference of FPR between group A and group B is: 
\begin{align}
\Lvert FPR_A - FPR_B \Rvert = 
\Lvert\mathbb{E}_{x,y,z}\left[ \frac{L_{A,-}}{\sum_{t=1}^{T}
\mathds{1}\{z=A\}\{y=-\} } - \frac{L_{B,-}}{\sum_{t=1}^{T}
\mathds{1}\{z=B\}\{y=-\}}\right]\Rvert
\end{align}

For the sake of notation we define 
$$C_{A,-} = \sum_{t=1}^{T}\mathds{1} \{y=-\}\mathds{1}\{z=A\}  \text{~~and~~}  
C_{B,-} = \sum_{t=1}^{T}\mathds{1}\{z=B\}\{y=-\}.
$$

Using Lemmas 1 and 2, we have:

\begin{multline}\label{eq:fairness1}
    \Lvert\mathbb{E}_{x,y,z}\left[ \frac{L_{A,-}}{C_{A,-}} - \frac{L_{B,-}}{C_{B,-}}\right]\Rvert \\
    \leq \Lvert\mathbb{E}_{x,y,z}\left[ \frac{(1+\eta)L_{f^*(A,-),A,-}}{C_{A,-}} +  \frac{\frac{\ln d}{\eta} }{C_{A,-}} + \frac{\sum_{t}q_{A,-}^t \cdot \alpha^t_{A,-}  }{C_{A,-}}  -  \frac{\gamma(\eta)L_{f^*(B,-),B,-}}{C_{B,-}} -  \frac{ \sum_{t}q_{B,-}^t \cdot\alpha^t_{B,-} }{C_{B,-}} \right] \Rvert \\
    = | (1+\eta) \mathbb{E}_{x,y,z}\left[ \frac{L_{f^*(A,-),A,-}}{C_{A,-}}\right]  - \gamma(\eta) \mathbb{E}_{x,y,z}\left[ \frac{L_{f^*(B,-),B,-}}{C_{B,-}}\right]  +  \\
    \left( \sum_{t}\frac{q_{A,-}^t \cdot \alpha^t_{A,-}  }{C_{A,-}} -  \frac{\sum_{t}q_{B,-}^t \cdot \alpha^t_{B,-}}{C_{B,-}} \right)  +
    \mathbb{E}_{x,y,z}\left[ \frac{\ln d }{\eta C_{A,-}}\right]|
\end{multline}

Using equation \ref{eq:epfair}, we have:
\[
\Lvert \mathbb{E}_{x,y,z}\left[\frac{L_{f^*(B,-),A,-}}{C_{A,-}}\right]  - \mathbb{E}_{x,y,z}\left[\frac{L_{f^*(B,-),B,-}}{C_{B,-}}\right] \Rvert \leq \epsilon
\]
Moreover, without loss of generality we assume that $f^*$ makes the smallest average loss on group B. This is,
\[
\mathbb{E}_{x,y,z}\left[\frac{L_{f^*(A,-),A,-}}{C_{A,-}}\right] \leq
\mathbb{E}_{x,y,z}\left[\frac{L_{f^*(B,-),A,-}}{C_{A,-}}\right]  \leq \mathbb{E}_{x,y,z}\left[\frac{L_{f^*(B,-),B,-}}{C_{B,-}}\right] + \epsilon
.\]

Thus, equation \ref{eq:fairness1} becomes:
    %
    \begin{align*}
    &\leq  \left\vert (1+\eta) \left(\mathbb{E}_{x,y,z}\left[\frac{L_{f^*(B,-),B,-}}{C_{B,-}}\right] + \epsilon \right)  - \gamma(\eta) \mathbb{E}_{x,y,z}\left[ \frac{L_{f^*(B,-),B,-}}{C_{B,-}}\right] 
    +
    \left( \frac{\sum_{t}q_{A,-}^t \cdot \alpha^t_{A,-}  }{C_{A,-}} -  \frac{ \sum_{t}q_{B,-}^t \cdot\alpha^t_{B,-}}{C_{B,-}} \right) +
    \mathbb{E}_{x,y,z}\left[ \frac{\ln d }{\eta C_{A,-}}\right]\right\vert \\
    %
     &\leq  \abs*{ \left( 1+\eta - \gamma(\eta) \right)   \mathbb{E}_{x,y,z}\left[ \frac{L_{f^*(B,-),B,-}}{C_{B,-}}\right] + \epsilon(1+\eta) + 
    \left( \frac{ \sum_{t}q_{A,-}^t \cdot\alpha^t_{A,-}  }{C_{A,-}} -  \frac{ \sum_{t}q_{B,-}^t \cdot\alpha^t_{B,-}}{C_{B,-}} \right) +
    \mathbb{E}_{x,y,z}\left[ \frac{\ln d }{\eta C_{A,-}}\right] }\\
    &\leq \Lvert\left( 1+\eta - \gamma(\eta) \right)   \mathbb{E}_{x,y,z}\left[ \frac{L_{f^*(B,-),B,-}}{C_{B,-}}\right] + \epsilon(1+\eta) +
    \left( \frac{ \sum_{t}q_{A,-}^t \cdot\alpha^t_{A,-}  }{p_A(1 -  \mu_{A,+})T} -  \frac{ \sum_{t}q_{B,-}^t \cdot\alpha^t_{B,-}}{p_B(1 -  \mu_{B,+})T} \right) +
    \frac{\ln d}{\eta p(1 -  \mu_{A,+})T} \Rvert \\
     &\leq \Lvert\left( 1+\eta - \gamma(\eta) \right) FPR_{f^*} + \epsilon(1+\eta) +
    \left( \frac{ \sum_{t}q_{A,-}^t \cdot\alpha^t_{A,-}  }{p_A(1 -  \mu_{A,+})T} -  \frac{ \sum_{t}q_{B,-}^t \cdot\alpha^t_{B,-}}{p_B(1 -  \mu_{B,+})T} \right) +
    \frac{\ln d}{\eta p(1 -  \mu_{A,+})T} \Rvert \\
     &\leq \Lvert\left( 1+\eta - \gamma(\eta) \right) FPR_{f^*} + \epsilon(1+\eta) +
    \left( \frac{ \sum_{t}q_{A,-}^t \cdot\alpha^t_{A,-}  }{p_A(1 -  \mu_{A,+})T} -  \frac{ \sum_{t}q_{B,-}^t \cdot\alpha^t_{B,-}}{p_B(1 -  \mu_{B,+})T} \right) \Rvert 
\end{align*}
where $FPR_{f^*}$ is the FPR of the best classifier $f^*$ on the best sensitive group $z^*$.
In the fourth line, when $T \to \infty$, the inequality follows from the fact that the last term goes to zero since its numerator is a constant . 

Let $q_{A,-}$ and $q_{B,-}$ indicates the converged true   value $q_{A,-}^t$ and $q_{B,-}^t$ respectively, where $q_{A,-} = \lim_{t \to \infty} q_{A,-}^t$ and $q_{B,-} = \lim_{t \to \infty} q_{B,-}^t$. Let $\delta_{A,-}^t = q_{A,-}^t-q_{A,-}$ and $\delta_{B,-}^t = q_{B,-}^t-q_{B,-}$ be the estimation errors at round t. By the classical central limit theory, the estimation errors converge at the rate of $O(\frac{1}{\sqrt{t}})$. Therefore,
\begin{align*}
    \frac{ \sum_{t}q_{A,-}^t \cdot\alpha^t_{A,-}  }{p_A(1 -  \mu_{A,+})T} -  \frac{ \sum_{t}q_{B,-}^t \cdot\alpha^t_{B,-}}{p_B(1 -  \mu_{B,+})T} &= \frac{ \sum_{t}(q_{A,-}^t-q_{A,-}) \cdot\alpha^t_{A,-}  }{p_A(1 -  \mu_{A,+})T} -  \frac{ \sum_{t}(q_{B,-}^t-q_{B,-}) \cdot\alpha^t_{B,-}}{p_B(1 -  \mu_{B,+})T}+ \underbrace{\frac{ q_{A,-}\sum_{t} \cdot\alpha^t_{A,-}  }{p_A(1 -  \mu_{A,+})T} -  \frac{ q_{B,-}\sum_{t} \cdot\alpha^t_{B,-}}{p_B(1 -  \mu_{B,+})T}}_{Q_{FPR}} \\
    &= \frac{ \sum_{t}\delta_{A,-}^t \cdot\alpha^t_{A,-}  }{p_A(1 -  \mu_{A,+})T} -  \frac{ \sum_{t}\delta_{B,-}^t \cdot\alpha^t_{B,-}}{p_B(1 -  \mu_{B,+})T}+Q_{FPR} \\
    &= \frac{ \sum_{t}O(\frac{1}{\sqrt{t}}) }{p_A(1 -  \mu_{A,+})T} -  \frac{ \sum_{t}O(\frac{1}{\sqrt{t}})}{p_B(1 -  \mu_{B,+})T}+ Q_{FPR}
\end{align*}
where the last inequality follows the fact that $\alpha_{A,-}^t<1$. Since
$\frac{ \sum_{t}O(\frac{1}{\sqrt{t}}) }{p_A(1 -  \mu_{A,+})T} < \frac{ O(2\sqrt{T})}{p_A(1 -  \mu_{A,+})T} $. When $T \to \infty$,  the estimation errors are sub-linear and thus go to zero.
Therefore,
\begin{multline}\label{eq:fpr_bound}
 \Lvert\mathbb{E}_{x,y,z}\left[ \frac{L_{A,-}}{C_{A,-}} - \frac{L_{B,-}}{C_{B,-}}\right]\Rvert 
\leq \Lvert\left( 1+\eta - \gamma(\eta) \right) FPR_{f^*} + \epsilon(1+\eta) +
\underbrace{\left( \frac{ q_{A,-} \cdot\sum_{t}\alpha^t_{A,-}  }{p_A(1 -  \mu_{A,+})T} -  \frac{ q_{B,-} \cdot\sum_{t}\alpha^t_{B,-}}{p_B(1 -  \mu_{B,+})T} \right)}_{Q_{FPR}}   \Rvert \\
\end{multline}
Similarly, for the absolute difference of FNR between group A and group B, we have:
\begin{multline}\label{eq:fnr_bound}
 \Lvert\mathbb{E}_{x,y,z}\left[ \frac{L_{A,+}}{C_{A,+}} - \frac{L_{B,+}}{C_{B,+}}\right]\Rvert 
\leq \Lvert\left( 1+\eta - \gamma(\eta) \right)FNR_{f^*} + \epsilon(1+\eta) +\underbrace{\left( \frac{ q_{A,+} \cdot\sum_{t}\alpha^t_{A,+}  }{p_A\mu_{A,+}T} -  \frac{ q_{B,+} \cdot\sum_{t}\alpha^t_{B,+}}{p_B\mu_{B,+}T} \right)}_{Q_{FNR}}   \Rvert
\end{multline}
where $FPR_{f^*}(FNR_{f^*})$ is the FPR(FNR) of the best classifier $f^*$ on the best sensitive group $z^*$.

\newpage

\begin{landscape}

\begin{table}[htbp]\caption{Table of Notation}
\centering 
\begin{tabular}{r c p{10cm} }
\toprule
$\mathcal{F}$   &   $\{f_1,\ldots,f_d\}$            &   Set of Classifiers\tabularnewline
$Z$             &                                   &   Random Variable for protected group \tabularnewline
$z$             &                                   &   Realization of $Z$: $z = A,z= B$\tabularnewline
$\mu_{z,y}$     &   $\mathbb{P}(Y=y|Z=z)$           & Base rate for group $Z=z$ and label $Y=y$\tabularnewline
$p_z$             &   $\mathbb{P}(Z=z)$               & Base rate for group $Z=A$ \tabularnewline
$\hat{y}$       &   $f(x,z)$                        & Prediction of a Classifier $f$ for input pair $(x,z)$ \tabularnewline
$w_{f,z,y}$     &                                   & Weight associated to Classifier $f$ for Group $z$ and Label $y$\tabularnewline
$\pi^t_{f,z,y}$  &   $\displaystyle  \frac{w^t_{f,z,y}}{\sum_{f \in \mathcal{F}}w^t_{f,z,y}}$        & Probability of selecting $f$ with weights $w_{f,z,y}$  at Round $t$\tabularnewline
$q_{z,y}$       &       & Probability of selecting $\pi^t_{z,y}$  \tabularnewline

$\pi_{f,z}^t$  & $\displaystyle  q_{z,+} \frac{w_{f,z,+}^t}{\sum\limits_{f \in \mathcal{F}}w_{f,z,+}^t} + q_{z,-} \frac{w_{f,z,-}^t}{\sum\limits_{f \in \mathcal{F}}w_{f,z,+}^t} $    & Probability of selecting $f$ at Round $t$ on Group $z$ (\model)\tabularnewline

$\ell_{f,z,y}^t$      &  $\ell_f^t \mathds{1} \{Z=z\}\mathds{1}\{Y=y\}$    & Loss of Classifier $f$ at Round $t$ on Group $z$ and Label $y$\tabularnewline
$\ell_{f,z}^t$      &  $\ell_f^t\mathds{1}\{Z=z\}$    & Loss of Classifier $f$ at Round $t$ on Group $z$ \tabularnewline
$\ell_f^t$      &  $\ell(\hat{y}^t_{f},y^t)$    & Loss of Classifier $f$ at Round $t$\tabularnewline
$L_{f,z,y}$ & $\displaystyle  \sum_{t=1}^T \ell^t_{f,z,y}$ & Cumulative loss of Classifier $f$ at Round $t$ on Group $z$ and Label $y$\tabularnewline
$L_{f,z}$ & $\displaystyle  \sum_{t=1}^T \ell^t_{f,z}$ & Cumulative loss of Classifier $f$ at Round $t$ on Group $z$\tabularnewline

$\mathbb{E}[ \ell^t_{z,y} ]$    &  $\displaystyle  \sum_{f \in \mathcal{F}} \pi_{f,z}^t \cdot \ell_f^t \mathds{1}\{Z=z\}\mathds{1}\{Y=y\}$ & Expected loss on Group $z$ and Label $y$ at Round $t$\tabularnewline
$\mathbb{E}[L_{z,+}]$ & $\displaystyle \sum_{t=1}^{T} \left( q_{z,+}^t  \sum_{f \in \mathcal{F}} \pi^t_{f,z,+}  \ell_{f,z,+}^t  + q_{z,-}^t  \sum_{f \in \mathcal{F}} \pi^t_{f,z,-}  \ell_{f,z,+}^t \right)$ & Expected cumulative loss on Group $z$ and Label $Y=+$\tabularnewline

$\mathbb{E}[L_{z}]$ & $\mathbb{E}[L_{z,+}] + \mathbb{E}[L_{z,-}]$ &  Expected cumulative loss on Group $z$ \tabularnewline

$\mathbb{E}[L]$ & $\displaystyle \sum_{z \in Z} \mathbb{E}[L_{z,+}] + \mathbb{E}[L_{z,-}]$ &  Expected cumulative loss \tabularnewline

$\alpha^{t}_{z,-}$ & $\sum_{f \in \mathcal{F}} \pi^t_{f,z,-} \ell_{f,z,+}^t - \sum_{f \in \mathcal{F}} \pi^t_{f,z,+} \ell_{f,z,+}^t$ & Expected difference loss using wrong $\pi^t_{f,z,-}$ and right $\pi^t_{f,z,-}$ probabilities \tabularnewline

$C_{z,y}$ &  & Cardinality of Group $z$ and Label $y$ \tabularnewline

\bottomrule
\end{tabular}
\label{tab:TableOfNotationForMyResearch}
\end{table}

\end{landscape}

\end{document}